# Confidence is the key: how conformal prediction enhances the generative design of permeable peptides

Laura van Weesep[1,2,3*] Sunay Chankeshwara[4], Leonardo De Maria[4], Florian David[5], Ola Engkvist[1,6], Gökçe Geylan[1,5*]

1: Molecular AI, Discovery Sciences, Biopharmaceuticals R&D, AstraZeneca, Mölndal, Sweden.
2: Department of Information Technology, Uppsala University, Sweden
3: Institute for Complex Molecular Systems (ICMS), Eindhoven University of Technology, The Netherlands.
4: Hit Discovery, Discovery Sciences, Biopharmaceuticals R&D, AstraZeneca, Mölndal, Sweden.
5: Department of Life Sciences, Chalmers University of Technology, SE-41296 Gothenburg, Sweden.
6: Department of Computer Science and Engineering, Chalmers University of Technology & University of Gothenburg, Gothenburg, Sweden
*email: [lauradesire.vanweesep, gokce.geylan]@astrazeneca.com

## ABSTRACT

Generative models coupled with reinforcement learning (RL), such as REINVENT and PepINVENT, have emerged as a powerful framework for *de novo* molecular design. During the ideation process these generative frameworks utilize various predictive models as part of the optimization objectives. However, the utility of the predictive models can be limited by their domain of applicability. When RL is used to explore the chemical space with predictive models, it can suggest molecules that lie outside the predictor's domain of applicability. As a result, the predictions may become less reliable, potentially steering designs into high reward but also high uncertainty chemical spaces. This is particularly pronounced for cyclic peptides which show therapeutic promise due to their modifiability and large interaction surfaces but are understudied compared to small molecules. While passive membrane permeation in cyclic peptides has attracted interest, identifying optimal permeable designs remains challenging yet crucial for targeting intracellular sites. We present an RL-guided generative framework that designs permeable cyclic peptides using an uncertainty-aware permeability predictor as the scoring component. To address predictive uncertainty, especially impacted by novel chemistry, we integrate conformal prediction (CP) as our uncertainty quantification method. CP assesses designs based on the calibrated model under a user-defined confidence level. We demonstrate that rewarding generated peptides with CP-informed predictions improves both reliability and efficiency of peptide optimization process. This also discourages exploration outside the predictor's applicability domain. This approach bridges the gap between predictive uncertainty and RL-guided exploration, showing how generative modelling and conformal prediction can be combined. It a practical framework for uncertainty-aware molecular design that efficiently navigates chemical space with quantified confidence.

## 1. INTRODUCTION

The therapeutic potential of peptides is gaining attention due to the potential advantages they offer over other modalities like small molecules[1,2]. These include their modifiability[3,4] and larger interaction surfaces, which can lead to higher affinity and specificity when targeting larger or more shallow surfaces[2,5]. Even though peptides are promising in targeting previously undruggable targets[4], their biological applicability to intracellular targets is severely limited by metabolic stability and cell permeability[6]. One approach to improve metabolic stability is incorporating natural amino acid analogues and peptide cyclisation[2], which could also improve peptide permeability[7]. Nevertheless, particularly, larger cyclic peptides, consisting of more than 10 residues, often pose a significant permeability challenge[8]. Despite ongoing computational efforts to decipher the patterns that enhance permeability, translating the insights into rule-based design criteria remains a challenge due to the intricate nature of this modality.

Machine learning (ML) has emerged as a tool that utilizes algorithms to learn patterns from data, with recently also offering advancements on peptide-focused applications[9,10], expanding beyond its traditional focus on small molecules[11,12]. Given the cost of peptide synthesis and technical demand of biochemical assays to assess peptide permeability, ML offers a valuable approach to accelerate the cell penetrating peptide design by predicting permeability. The CycPeptMPDB[8] database, compiled from over 45 publications with over 6000 peptides in total, has enabled a more systematic evaluation of predictive models for permeability. A range of approaches have been developed to predict permeability[13], including support vector machines, random forests, and more recent graph neural networks like multi_CycGT[14] and multi-assay models[15]. A recent study further examined the challenge of predicting permeability, looking at the generalizability of algorithms beyond their applicability domain. Using a conformal prediction framework, the authors showed that, although model performance was maintained within the applicability domain defined by the training data, reliability deteriorated in more distant regions of chemical space and could be partially restored through recalibration with a subset of data from the target domain[16]. Nevertheless, fundamental questions remain about model reliability when predicting novel peptides that differ significantly from training data, a critical concern when using these models to guide generative design.

Reinforcement learning (RL) has been adopted in many *de novo* drug design applications[17–20] as it can be utilized for generating molecules with specific desirable properties. RL for *de novo* design typically involves the use of a generative model and various scoring components such as calculated properties or predictive models. While the generative model facilitates the exploration of the molecular design space, the scoring components play a critical role in steering the model towards desired outcomes by refining the search space. Feedback based on the scores of the generated molecules is then utilized to iteratively improve the new set of generated molecules, enhancing its ability to produce more effective drug candidates. REINVENT is one of the RL-based frameworks that allows for straightforward integration of scoring components, offering different flavors addressing a range of applications[21]. More recently, also PepINVENT has become available which allows RL-based design for peptidic molecules[22]. PepINVENT[23] proposes natural and non-natural amino acids to fill in query positions of an input peptide sequence. In each optimization step, the generated amino acids are mapped to query positions, and the completed peptide is scored in the feedback loop (*Figure 1*).

Predictive models for peptides are commonly used as a scoring component in generative tasks[22,24,25]. A main challenge when using predictive models in the scoring function of RL is quantifying how reliable these models can predict on new instances outside the training data. The model probabilities are usually seen as an indication of “uncertainty”, it is however not evident how reliable this "uncertainty” is, plus the calibration of these uncertainties is often missing. Consequently, predictive models used in scoring functions may produce unreliable predictions for generated molecules outside their applicability domain. This can mislead the RL agent, waste computational resources or, worse, generate molecules with falsely confident predictions.

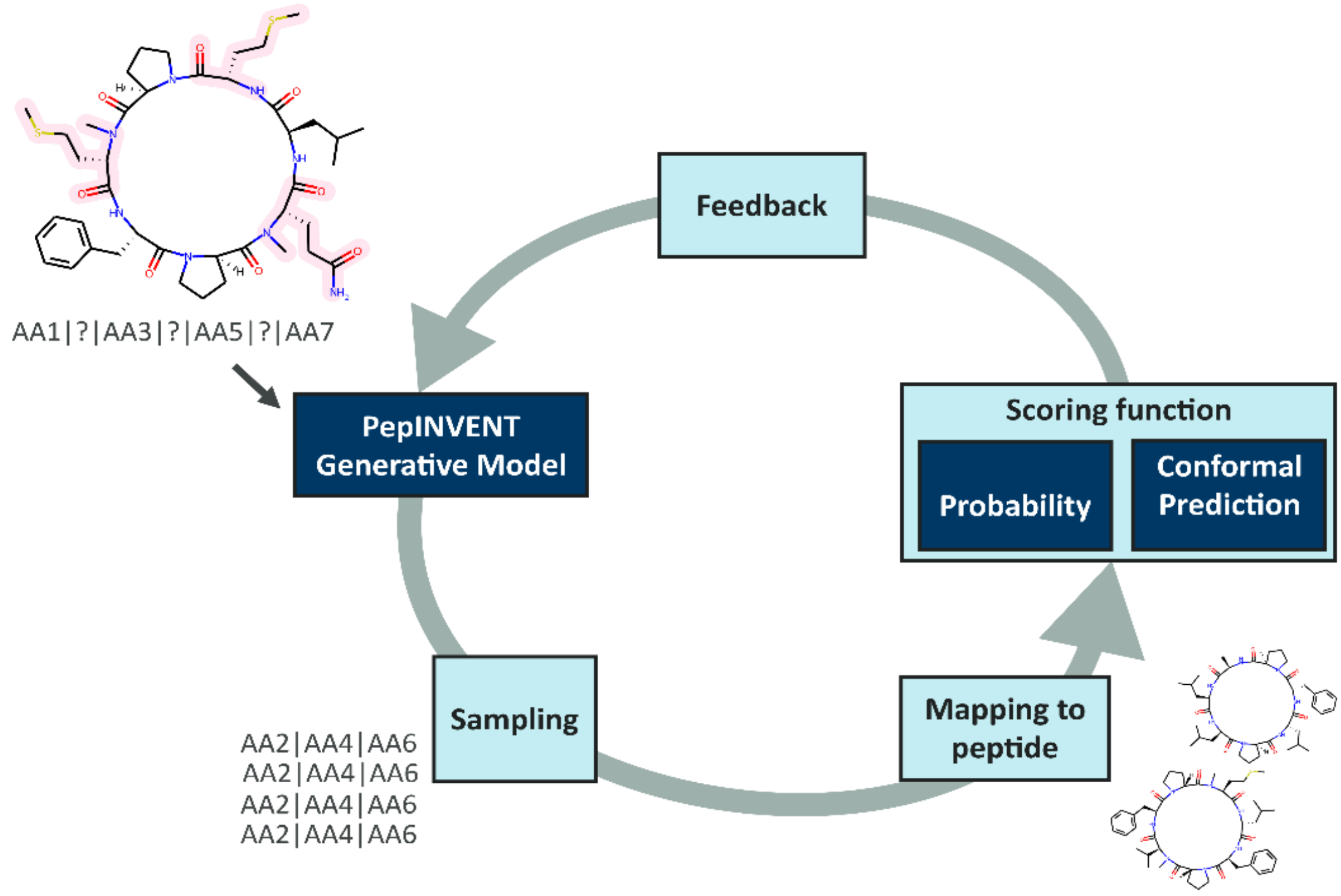

*Figure 1: Schematic overview of the generative design in the reinforcement learning loop. A macrocyclic peptide with multiple amino acids intended for modification, input to the generative model. The generative model produces CHUCKLES-based SMILES for the masked amino acid positions. Proposed amino acids for these positions are subsequently mapped into the full peptide to assemble the generated peptide. This peptide is then scored using the scoring function. Based on the assigned score, feedback is provided to the generative model, which iteratively proposed new amino acids to optimize for higher scores.*

When applying a model to predict new data, it is recommended to assess the model's applicability domain[26,27], defined as the chemical space and response value range in which the model makes reliable predictions[28]. When the new molecule is similar to the training data and falls within the applicability domain, the model's predictions are expected to be more reliable. In contrast, if the molecule is outside this domain or dissimilar, the model may yield predictions that lack sufficient information, resulting in reduced reliability. Conformal prediction is one of the methods to approximate this reliability.

Conformal prediction (CP) is an uncertainty quantification framework that is implemented on top of traditional ML algorithms. The framework calibrates predictions producing valid confidence estimates specific to individual predictions[29]. Scores reflect how closely the predictions resemble those made on the data used to train or calibrate the ML model. The fundamental assumption of CP is that test examples are independent and drawn from the same distribution as the calibration set; an assumption frequently made when employing QSAR models[30]. Mondrian inductive conformal prediction (ICP) enables class-specific uncertainty estimation, providing calibrated confidence scores independently for each class. In binary classification, the model can give one of the 4 predictions: i-ii) single prediction classes, such as "Class 0" and "Class 1", iii) "None" class indicating that the predicted compound is outside of the applicability domain, and iv) "Both" class when there is a lack of information to distinguish between the two classes. These predictions are decided under the user-specified significance level. To illustrate, an 80% confidence level corresponds to a significance level of 0.2, implying predictions are valid if the error rate is below 20%. Here, an error occurs when a prediction does not include the correct label, with "Both" always being accurate. To avoid the model exclusively outputting "Both", efficiency serves as a secondary metric. Efficiency tracks the fraction of predicting one of the single classes. The significance level impacts the trade-off between efficiency and validity. Furthermore, to improve conformal efficiency, usually multiple ICPs are aggregated to construct an aggregated conformal predictor (ACP)[31]. Beyond performance guarantees under the exchangeability assumption, ICPs can recalibrate predictions of the ML models to a related domain, by adjusting the calibration set. This allows for recalibration without the need for retraining the underlying model, when predicting on datasets distinct from the training data[32]. These qualities make CP a promising approach for assessing uncertainty in predictive models.

This study aims to leverage elements of conformal prediction and reinforcement learning within the context of *de novo* design of cyclic peptidic molecules, focusing particularly on optimizing cell permeability. By integrating CP directly into the RL loop, we not only aim to enhance the reliability of the sets of molecules generated, but also the efficiency of exploration. Our study addresses the crucial aspect of connecting the uncertainty inherent in predictive models with the strategic exploration and learning behavior dictated by RL, aiming to enhance the effective use of predictive models within this context. By doing so, we aim to create a robust framework that navigates the vast chemical space with a higher degree of certainty and strategic foresight.

# 2. RESULTS

In the following sections, we will explore the effectiveness of CP as a scoring component in RL, with a focus on improving passive peptide permeability. Our objective is to identify peptides that are predicted to be permeable with high confidence. In this work, the quality of the method is measured by monitoring the efficiency of accessing high-reward spaces and validity of the generated samples. Efficiency, in this context, refers to the number of learning steps to reach either an average fraction of 0.5 confident permeable predictions or a plateau. Validity is defined as the fraction of syntactically valid SMILES strings when generated amino acids are mapped back to the query positions of the input sequence.

We evaluated the impact of CP within the RL framework across a test set of 150 queries, running each query independently. The queries consisted of 6- to 12-amino acids long cyclic peptidic molecules from Bhardwaj et al.[33], hereafter referred to as the Baker dataset. Up to 4 amino acids of the test set peptides were randomly masked, tasking the generative model to generate the missing amino acids.

Following PepINVENT's framework, generated peptides were scored for permeability with an XGBoost classifier (*see Methods 5.2*). The permeability was assessed using multiple evaluation schemes that vary in how predictive confidence is treated[23]. For each RL run, CP-based P-values of $P_1$ (the permeable class) and $P_0$ (the non-permeable class) were tracked alongside the scoring scheme of interest. The P-values enabled monitoring confidence in the assigned permeable and non-permeable classes. $P_1$ above and $P_0$ below the significance level of 0.2 indicated a confident, also conformally efficient, prediction of permeable class.

## 2.1. Using Uncertainty-unaware Model Predictions

Classifiers in RL are typically employed by propagating model probability of the desired class[34]. Therefore, the model probabilities of the “permeable” class, or class 1, from the XGBoost classifier was used as our baseline. Since there are no uncertainty estimates or score transformations applied on these predicted probabilities, we refer to these as raw model scores.

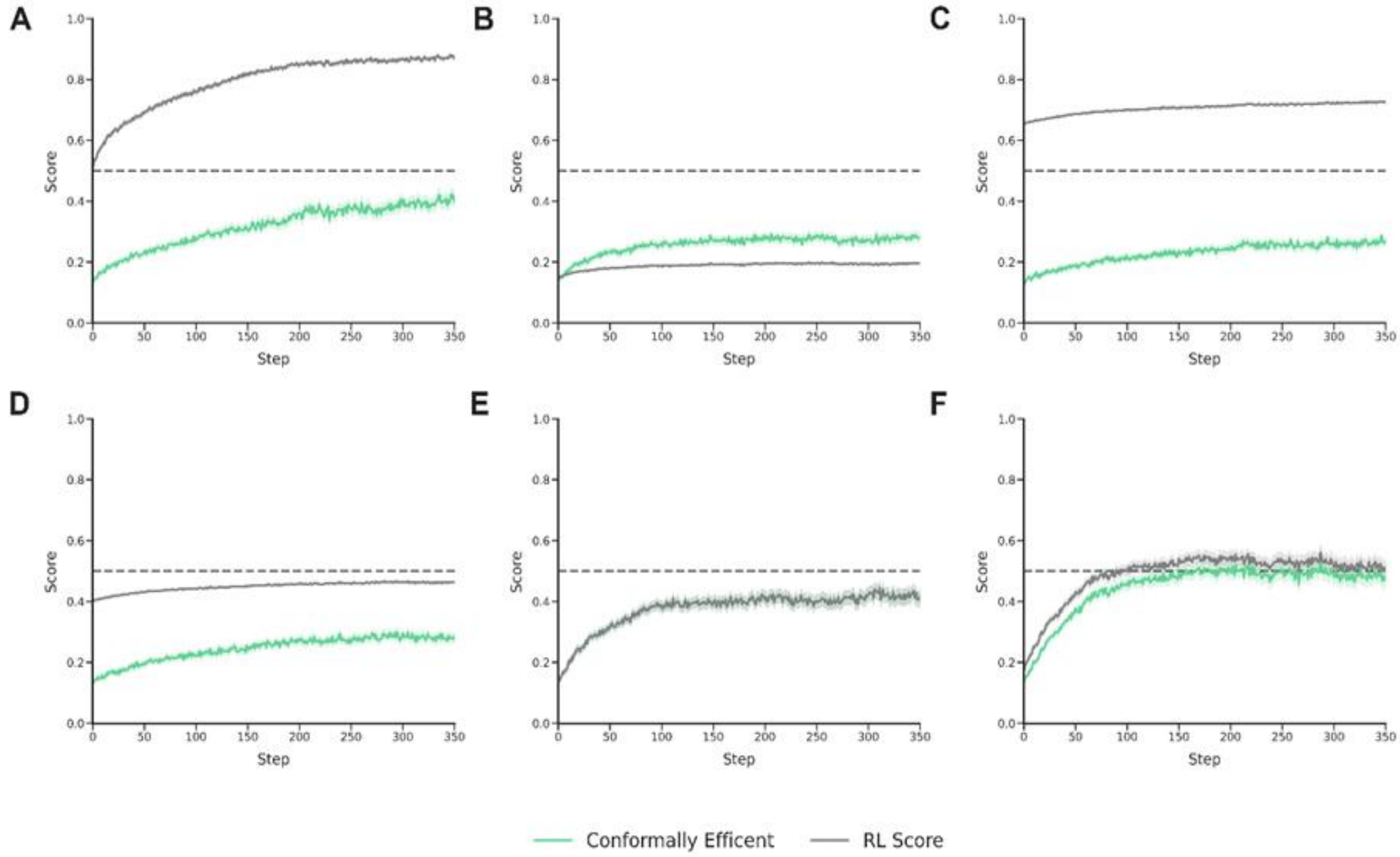

*Figure 2: Learning progress during RL run with the investigated scoring functions for permeability objective. RL processes are shown in score versus learning step plots. The scoring functions explored were (A) the probability of class 1 based on the raw model representing the baseline, and (B-F) various scoring functions built to capture the model uncertainty from the CP model. The CP-based functions include (B) $P_1$, (C) 1-$P_0$, (D) $P_1$-$P_0$, (E) harsh function, and (F) soft function. The plots include RL scores (grey) referring to the output of the evaluated scoring function and the fraction of conformally efficient molecules predicted to be permeable (green). Conformally efficient predictions are the molecules that fulfill $P_1$>0.2 and $P_0$<0.2. Evolutions of each scoring component over 350 steps for the unique valid molecules, are shown as both the average score and the span across the generated peptides for 149 independent test cases.*

During the RL runs for 150 test molecules, the average RL score, or raw model score, of the valid molecules in the batches were calculated. The raw model scores increased from 0.51 to 0.87 over 350 epochs (*Figure 2A*). This suggests that the generative model learns to propose peptides with an increased probability of being classified as permeable. To determine whether the rise in predictive probability corresponds to a similar increase in confident permeable predictions, we also classified the generated peptides with conformal prediction. It is expected that as raw model probabilities increase, the $P_1$ score will increase and $P_0$ score will decrease, leading to more permeable predictions that are also confident. The results agreed with this expectation as the number of conformally efficient predictions for permeable class increased with the increased RL score (*Figure 2A*). This was due to $P_1$ increasing from 0.15 to 0.23 and $P_0$ decreasing from 0.35 to 0.23, on average (*Supplementary Figure 1A*). However, the number of conformally efficient predictions remained much lower against the raw model's "permeable" predictions. This highlighted the gap between high scoring peptides and confident predictions. One of the main reasons for confidence is P0 remaining over 0.2 throughout the run, being the fundamental bottleneck for our model. The model cannot confidently distinguish permeable from non-permeable peptides for the generated molecules. This demonstrates that high raw model probabilities do not guarantee reliable predictions, highlighting the need for uncertainty-aware scoring.

## 2.2. Conformal Prediction in the Scoring Function

Following the limitations of the raw model, we investigated the ability of RL to generate permeable peptides with high confidence predictions. We explored different ways of incorporating conformal prediction directly in the scoring function. Since CP-based P-values are calculated independently, incorporating one of them alone was considered to mimic scoring with raw model probabilities. The two scoring objectives considered were maximizing $P_1$ and (1- $P_0$) values from the conformal predictor. The rationale is to increase conformal evidence for permeability by either maximizing $P_1$ directly or, equivalently, minimizing evidence for non-permeability by maximizing $(1 - P_0)$. Additionally, we explored three scoring functions combing $P_1$ and P0: i) $P_1$ - $P_0$ in which $P_1$ and 1- $P_0$ could synergistically drive the generation towards identifying more permeable molecules, ii) a "harsh" scoring function in which CP was assessed in a binary scheme only rewarding conformally efficient predictions with $P_0$ <0.2 and $P_1$ >0.2 and iii) a "soft" scoring function allows for a partial reward when either one of the two requirements for conformal efficiency is met.

The generative runs show increasing RL scores for all scoring functions utilizing CP. However, the single P-value scoring functions, $P_1$ and (1- $P_0$), showed marginal improvement of the average RL scores (Figure 2B-D). The limited learnability of the single P-value scoring functions indicates that adjusting the certainty of one of the two P-values does not necessarily guarantee confident predictions. Although, $P_1$ and (1- $P_0$) showed a continuous increase in the fraction of conformally efficient predictions, they do not reach the levels achieved by scoring with the raw model probability. On top of this, the confidence in the permeable label is limited (Supplementary Figure 1B, C). These results suggest that maximizing confidence in the permeable class alone is insufficient, we must also minimize likelihood of belonging to the non-permeable class. Decoupling the conformal P-values and using a single one as the scoring objective hinders learning the joint decision produced by Mondrian ICP.

Combining $P_0$ and $P_1$ addresses the limitations of using individual metrics by requiring high confidence in permeable and low confidence in non-permeable classes. However, a straightforward combination such as $P_1$ - $P_0$, led to similar results that suggested that stricter guidance on steering the generation towards chemical spaces with lower uncertainty may be necessary.

The "harsh" and "soft" functions were proposed to provide a direct integration of conformal efficiency where a confident single-label "permeable" class assessment is rewarded. Harsh scoring function showed a steep increase followed by a stable score over 350 epochs, indicating that it is effectively learnable (*Figure 2E*). As for the soft function, the generation reaches the target chemical space in the least number of steps, after only 150 epochs (*Figure 2F*). In this case, we defined the target chemical space with 50% of the valid molecules with confidently predicted as permeable. Therefore, we can conclude that the soft scoring function results in the fastest convergence as well as the greatest fraction of molecules that were confidently predicted as permeable, compared to all the scoring functions explored. It is worth highlighting that the batch averages of $P_0$ and $P_1$ do not converge to below and above the significance level of 0.2, to satisfy the conformally efficient predictions (Supplementary Figure 1). However, $P_1$ reached the highest average scores compared to other scoring methods. Since both harsh and soft scoring functions provide discreet reward schemes, the maximum reward is reached when $P_1$>0.2 and $P_0$<0.2 for a peptide. Even though raw model probabilities provide more peptides with confident permeable predictions. Therefore, we conclude that the soft scoring function is an effective approach to generate reliable permeable peptides. To further improve the confidence of predictions in practice, one would need to tune the significance level.

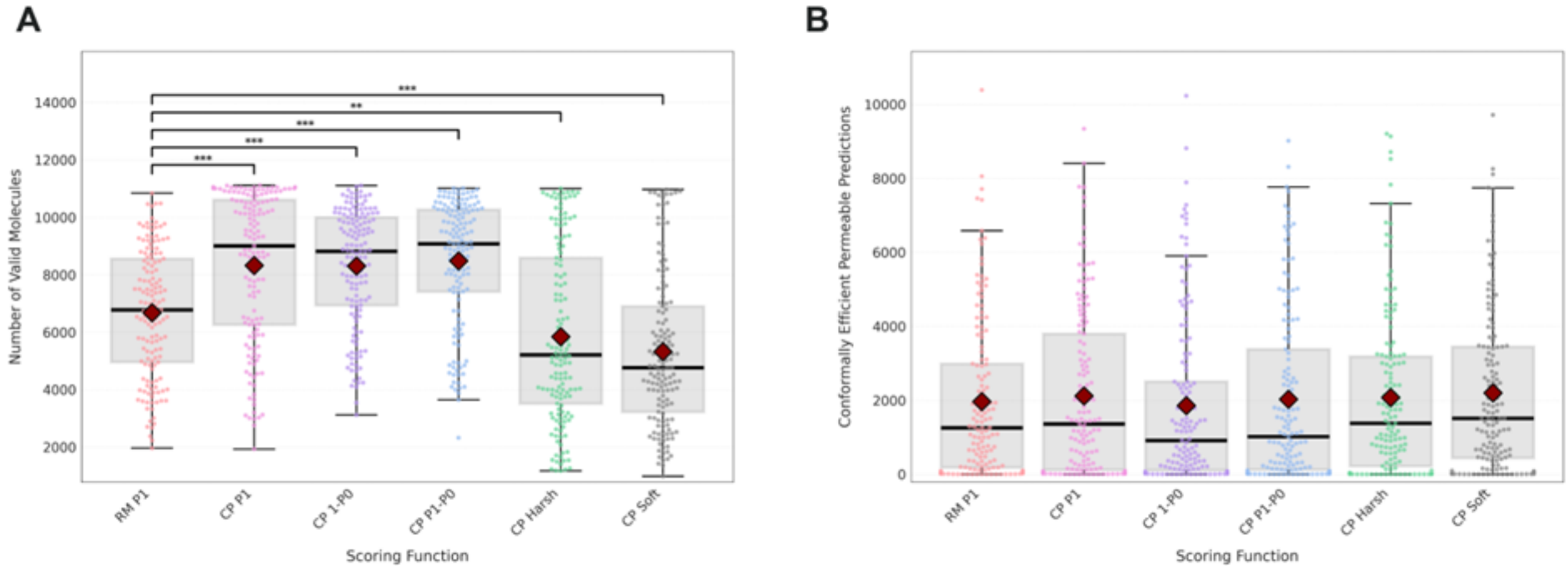

*Figure 3: The number of valid molecules generated (A) per scoring function and the number of peptides predicted to be permeable in a conformally efficient way (B). The red diamond corresponds to the mean over 149 cases, with each point corresponding to an individual test case. The number of valid scoring functions was compared to the RM p1 scoring function using a Wilcoxon test, with * corresponding to p<0.05, ** corresponding to p<0.01 and *** p<0.001. Despite the number of valid molecules being significantly different, no significant difference in the number of conformally efficient molecules was found.*

Motivated by the seeming discrepancy between $P_0$ and $P_1$ improvements for the raw model versus the CP soft scoring function and the improvement in conformally efficient permeable predicted molecules further investigation was done. We examined whether the apparent advantage of one scoring function over the other depends on how a "hit" is defined. On the left, we compare the number of peptides classified as permeable by the raw model probability for the RM scoring function and the soft scoring function. By this definition, the raw model scoring function appears superior, yielding more hits than the CP soft scoring function. On the right, we consider the number of conformally efficient permeable predictions, i.e., predictions that are both permeable and within the model's applicability domain according to conformal prediction. Under this criterion, the CP soft scoring function produces more peptides than the raw model. This indicates that optimizing for raw model probability inflates apparent hit rates without improving calibrated, domain-consistent permeability, while the CP soft scoring function better targets reliable, well-calibrated hits.

## 2.3. Impact of Scoring Functions on Chemical Validity

Chemical validity is one of the important aspects in evaluating the impact of scoring objectives in the generative process. On one hand, receiving no reward over a number of consecutive steps for the entire batch can steer the generation to producing invalid molecules. This is particularly plausible in sparse reward scenarios. On the other hand, mode collapse can occur when the generative model discovers peptides that receive high reward and generate them over and over again[34]. In such cases, validity of the unique molecules we obtain can be significantly reduced. 149 out of 150 cases, we did not encounter this. However, one of the molecules, test molecule 7, led to an error for the harsh scoring function, therefore got omitted from all the results. For the rest of the test cases, the number of unique valid molecules over 350 epochs was calculated.

The average number of unique valid molecules for the soft scoring function was lowest out of all the scoring functions tried (*Figure 4A*), but the mean number of conformally efficient permeable predictions was highest out of all scoring functions tried (*Figure 4B).* This highlights that looking at valid molecules only, does not give guarantees for the highest number of molecules meeting your criteria. On top of this, it was noted that the soft scoring function in the worst cases (lowest $10^{th}$ percentile), produces more valid molecules than the harsh function (2282.8 vs 2172.8), indicating fewer failures, likely related to the reward sparsity due to the discreet scoring. The reason for this might be that soft scoring function contains a partial reward leading to a more gradual steering toward the desired chemical space. The significantly lower number of valid molecules being generated whilst generating more conformally efficient permeable predicted molecules highlights that the RM p1 score does not efficiently exploit the space. It generates more molecules, but a higher ratio of those molecules does not meet the criteria of being conformally efficient permeable predictions. Consequently, the soft function was considered to be a robust scoring strategy. It effectively maximizes the difference between $P_1$ and $P_0$ while also maintaining sufficient high validity during peptide generation to not end up generating only invalid molecules. Other scoring functions such as the $P_1$, $(1-P_0)$ and $P_1-P_0$ scoring functions generated significantly more valid molecules compared to the RM (*Figure 2A)*, though the RL results and the number of conformally efficient permeable predictions uncover that many of these molecules are low scoring molecules (*Figure 1* and *Figure 2B*).

### 2.4. Effect of Peptide Length on Performance

As an additional check for robustness of the soft scoring function, we evaluated how the generation is impacted on the peptide length. Peptides of lengths 6, 7, and 10 dominate the predictor's training data.[8] This structural imbalance may influence confidence and shapes the applicability domain. Therefore, we investigated how each scoring function performs in RL on test cases with peptide lengths of 6 to 12. The results indicated that the soft scoring function can effectively learn the objective and deliver permeable designs with confidence for lengths 6, 7, 8 and 10 (Supplementary Figures 2-4, 6). In these cases, $P_1$ exceeds $P_0$ and reaches this separation in fewer epochs than the raw model (*Supplementary Figures 2-4, 6*). Whilst for lengths 9, 11 and 12, the performance of the soft scoring function deteriorates which would yield lower number of peptides with confident permeable labels (Supplementary Figures 5, 7, 8). The same behavior was observed for the fraction of conformally efficient permeable predictions. For lengths 6, 7 and 8, 50 epochs half of the valid peptides are predicted to be permeable in a conformally efficient way (*Supplementary Figures 2-4)*. For peptides of length 10 it takes roughly 75 steps (*Supplementary Figure 6)*, whilst for the peptides of length 9, 11 and 12 the point of 50% of the peptides being predicted to be permeable in a conformally efficient way is not reached within 350 epochs (*Supplementary Figures 5, 7 & 8)*. We also see that for all lengths, except for length 12, where not continuous learning was seen by the soft scoring function, the soft scoring function always reaches the highest fraction of conformally efficient predicted permeable peptides within less epochs than the raw model scoring function. This shows that having uncertainty in the design loop can help identify pitfalls of the predictors as scoring components. Also, it can inform the user when to stop relying on the proposed peptides in the generative design for a particular objective.

## 3. Discussion

This research demonstrates that integrating conformal prediction directly into RL-guided molecular design enables more reliable and efficient exploration of chemical space. It steers the generation toward high scoring regions with calibrated confidence. In this research, we investigated how this can be used for *de novo* design of permeable cyclic peptides. In PepINVENT, the generative model proposes a set of designs that is rewarded based on the predictions of the property predictor[22]. In this study, this property is cell permeability. The generative model fine-tunes its proposed molecules iteratively based on this reward feedback.

However, these models navigate vast chemical spaces and can generate molecules that are significantly different from the training data of the predictive model over learning steps. In such cases, the predictions reduce to random outputs as the molecules fall outside of their applicability domain. This in turn makes the generative process unreliable with objectives not fulfilled effectively. If we consider employing multiple predictive models in RL loop, the additive uncertainty could dramatically reduce the quality of and the trust towards the proposed molecules.

We hypothesized that integrating conformal prediction would steer RL toward regions of chemical space with higher-confidence permeability. Thus, we conducted virtual experiments to test whether scoring through uncertainty quantification in RL yields more permeable peptides than using raw model probabilities alone. Raw model probabilities represented our baseline as the uncertainty-unaware method and the current practice in generative design. Although raw model probabilities were readily learned by the RL policy, the resulting predictions were, on average, unreliable at the 80% confidence level. This indicated that the significant portion of the high-scoring permeable designs would be questionable according to this method.
Scoring in the RL loop under uncertainty was assessed through various scenarios. Conformal prediction, using Mondrian ICP, utilizes two independent P-values to yield a decision. In the scenarios explored, we observed that using a single P-value as the scoring component, maximizing $P_1$ or minimizing $P_0$ was effectively learned without sacrificing validity of the generated peptides. However, learning either objective alone did not increase the number of peptides confidently predicted to be permeable. Dismantling the duality of CP showed no guarantee of reciprocity: maximizing $P_1$ does not ensure low $P_0$, and minimizing $P_0$ does not ensure high $P_1$. When conformal efficiency itself was used as the scoring scheme, both P-values are utilized. soft scoring function formulated this through a discreet reward scheme with a partial reward when one of the P-values reached its objective. This guided RL to explore chemical space where the predictive model could make reliable distinctions between permeability classes at 80% confidence. Also, the soft scoring function with partial reward provided sufficient gradient for learning while generating more unique conformally efficient permeable predicted peptides compared to its binary counterpart, the harsh scoring functions. This approach kept peptide generation within the predictor's applicability domain, avoiding the common pitfall where RL optimizing into unreliable chemical space. Results highlighted not only a higher proportion of permeable molecules but also a faster convergence to the high reward space. Notably, even after the harsh and soft rewards reach a steadier slope and slowly plateau, still unique conformally efficient permeable predicted peptides are generated, indicating that the model is not constrained from further exploring or exploiting the chemical space.

The peptides evaluated here represent a small subset of the possible design space, therefore the optimal scoring function may differ for other predictive tasks. We also observed that the choice of query peptide influences learning performance. In particular, conformal efficiency can be harder to achieve when unmasked amino acids impose overly restrictive constraints or when too few positions are masked to shift the peptide's properties into the desired region. We also observed length-dependent performance variations. Optimization was more effective for peptide with lengths well-represented in training data (6, 7, and 10 amino acids). Conformal metrics reflected this by indicating these lengths were within the predictor's applicability domain, consistent with better model performance reported previously[13]. Thus, these peptides were effectively optimized with conformal scores. Future work could focus on investigating other nonconformity scores such as Tanimoto similarity that focuses on the input, instead of raw model probabilities thus decoupling underlying model probabilities from uncertainty.

Experimenting with other recalibration methods such as Platt-scaling or Venn-Abers[35], would be another interesting avenue to pursue. Also, integrating multiple calibrated predictors in the scoring function and investigating their cumulative impact on RL-guided generative process can yield new bottlenecks, especially when the applicability domains of predictors are significantly different from each other.

## 4. Conclusion

In conclusion, our approach addresses the crucial aspect of connecting the uncertainty inherent in predictive models with the strategic exploration and learning behavior dictated by RL. For the first time, we present an uncertainty-aware generative framework by integrating conformal prediction with reinforcement learning-guided generative design. We demonstrate that factoring user-demanded confidence directly in scoring navigates chemical space more efficiently to identify a greater number of peptide designs confidently predicted to be confident permeable. This confidence governs conformal efficiency and can be tuned to apply optimization pressure within the predictor's applicability domain.

Applied to cyclic peptide permeability, CP-based peptide scoring produced more suitable candidates and sustained 80% confidence for single-label permeable predictions. Among the strategies tested, the soft scoring function offered a smoother reward landscape that improved convergence speed to high reward spaces and maintained exploration within trustworthy regions of chemical space. These results underscore the broader potential of the synergistic effect between CP, RL, and generative models to enhance reliability and efficiency across molecular design tasks beyond peptide permeability.

## 5. METHODS

This methods section describes the RL framework for peptide design, the predictive model with CP for permeability prediction, and various strategies used to incorporate this predictor into the RL loop as a scoring component.

### 5.1. Reinforcement Learning Loop

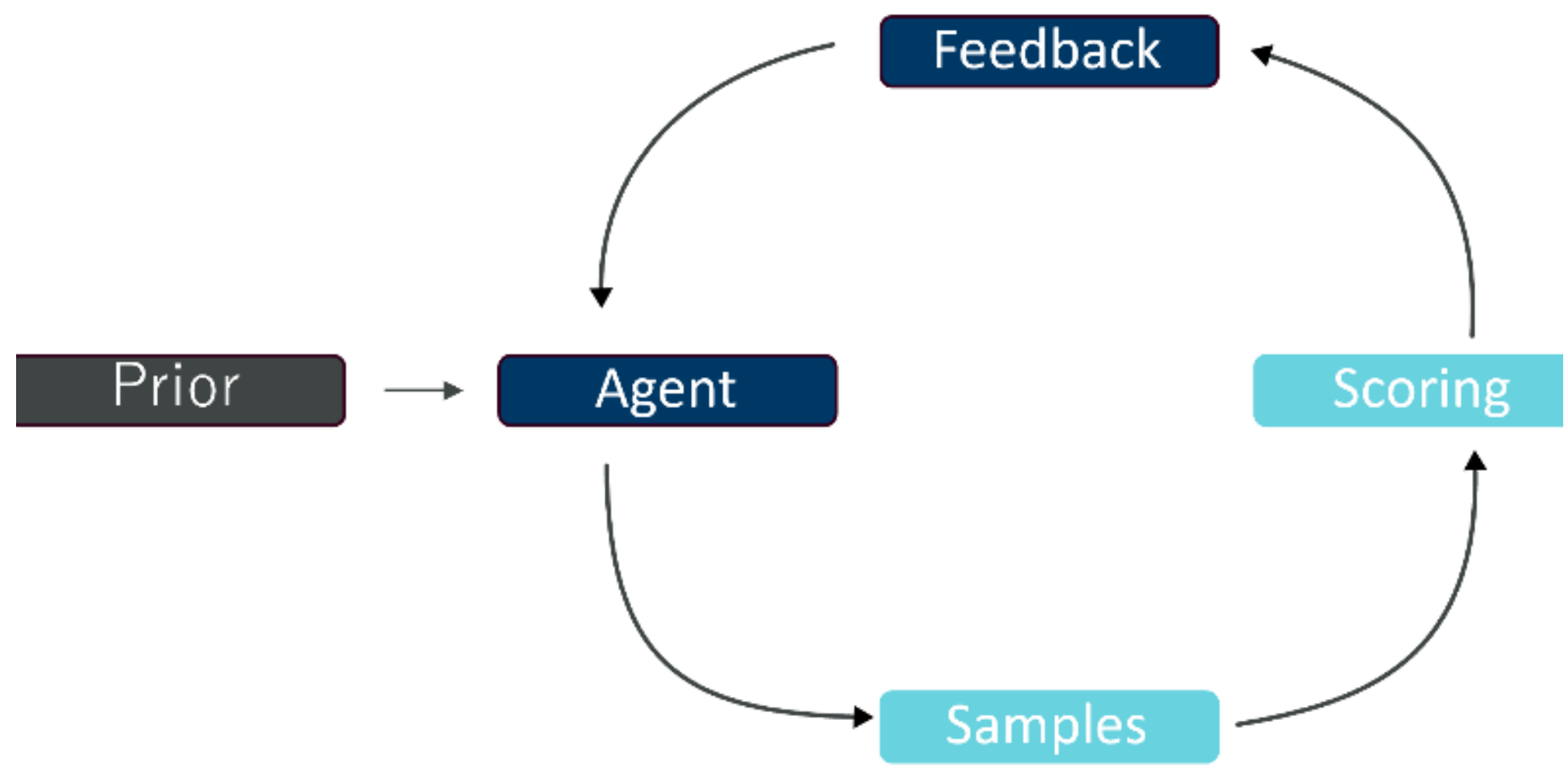

*Figure 5: Overview of an example of how RL can be applied to drug design*

For this research, we employed the PepINVENT framework. This model is a pretrained transformer trained on semi-synthetic data, with building blocks including the natural and a library of 10,000 non-natural amino acids[23]. To train the network, peptides were written in the following format: `aa1|aa2|aa3|aa4|aa5`, where each amino acid, aa, separated by "|" is written in the CHUCKLES notation[36]. CHUCKLES provides a chemical representation for peptides beyond the natural amino acids by using the simplified molecular input line system (SMILES) notation[37]. CHUCKLES writes amino acids in the peptide sequence in the standardized format of N-to-C and therefore allows them to be arranged sequentially from N-terminus (amino group) to C-terminus (carboxyl group) in the peptide chain[38]. The transformer model was trained on a masking task, in which random amino acids are masked, and the task was to propose amino acids to fill in these masked positions. This approach enables the model to generate molecules within the context of the entire peptide, including the knowledge of cyclisation, disulfides, as well as general amino acid characteristics.

A simplified overview of the RL framework can be seen in *Figure 5*. The RL loop makes use of two generative models, the prior and the agent. The generative models estimate the likelihood of producing specific tokens $t$ in a sequential matter (Equation 1) [21]. In the RL loop, generated molecules are scored, and feedback is given to narrow down the generative space to a desired region. During training, the scores are calculated for each learning step, and the likelihood score given by the prior (unbiased generative model), the agent (biased generative model), and the augmented likelihood are calculated by Equation 2. Based on these, a loss is calculated by equation 3, which feeds back information to the agent to generate different molecules[39].

The negative log likelihood (NLL) for the prior and the agent is calculated in the following way:

$$\mathrm{NLL}(\mathrm{T}|\mathrm{S}) = -\log \mathbf{P}(\mathrm{T}|\mathrm{S}) = -\sum_{i=1}^{l} logP(t_i|t_{i-1}, t_{i-2}, \ldots, t_1,\ S)^{21}, \qquad \text{(Equation 1)}$$

With the NLL of the sequence of characters *T,* given input sequence *S,* defined as the sum of the negative logarithmic probabilities ($logP$) over each character $t_i$ given the preceding characters $t_{i-1}, t_{i-2}, \ldots, t_1$ and input sequence S.

The augmented likelihood for each sequence is calculated by:

$$logP_{aug}(T) = \log P_{prior}(T) + \sigma S(T), \qquad \text{(Equation 2)}$$

where $logP_{aug}(T)$ is the augmented log likelihood providing a balance between exploration and exploitation. $logP_{prior}(T)$ is the likelihood calculated for a SMILES sequence based on the prior, serving as a metric to ensure the validity of the SMILES. $S$ is the score given by the scoring function and $\sigma$ is the weight given to the scoring function. In our case, we used a weight of 50.

The loss is calculated using the following formula:

$$\mathcal{L}(T) = \left(\log P_{aug}(T) - logP_{agent}(T)\right)^2 \qquad \text{(Equation 3)}$$

[21].

Based on the loss, the parameters of the generative model are adjusted to converge towards a space that minimizes the loss.

## 5.2. Predictive model for Permeability

### Dataset description

For training the permeability classifier, the Parallel Artificial Membrane Permeability Assay (PAMPA)[40] dataset from CycPeptMPDB was used[8]. This contains data from the PAMPA assays, which are used to quantitatively predict passive diffusion across phospholipid membranes. In total, this dataset contains experimental results from 47 different sources for cyclic peptides and peptidic macrocycles, made up of 312 different types of monomers.

6941 non-conjugated cyclic peptide entries with PAMPA assay results were downloaded and standardized with `RDKit v. 2022.03.2.22`[41]. After canonicalization with chirality, the value of $LogP_{exp}$ higher than or equal to -6.00, was used to differentiate between high and low permeability based on previous research[8]. The data processing was finalized after eliminating the duplicates and the peptides with conflicting labels due to measurement variations from different labs. The data processing yields 6876 data points, 2228 non-permeable peptides, and 4648 permeable peptides. with cycle sizes ranging between 12 to 46 atoms.

### XGBoost Model

Similar to previous work[22], a `XGBoost v2.0.3`[42] classifier to predict permeability was built. In this scenario, 10% of the data was allocated to a test set and 90% to the training set. Based on prior work establishing extended connectivity fingerprints (ECFP) as molecular descriptors giving a robust performance of peptide-related predictors[24,43,44]. ECFP with radius of 2, 2048 bits, `useCounts=True`, `useFeatures=False`, `useChirality=True` were used to encode the cyclic peptidic molecules[45]. Hyperparameters selected for the model based on 10-fold cross-validation using Bayesian search are displayed in *Table 1*. Model training and early stopping were conducted based on F1-score. The raw model achieved a balanced accuracy of 78% and a sensitivity of 90%.

*Table 1: Parameters used to train the XGBoost model for permeability classification*

| Hyperparameter | Value |
|---|---|
| Colsample by level | 0.30 |
| Colsample by tree | 0.35 |
| Learning rate | 0.31 |
| Max depth | 30 |
| N estimators | 295 |
| Subsample | 1 |

## Conformal prediction

For this research we used an Inductive Mondrian Conformal Prediction (ICP)[30]. The, CP provides two p-values: $P_0$ (confidence the molecule is non-permeable) and $P_1$ (confidence the molecule is permeable). A molecule is conformally efficient when one p-value is high (>0.2) and the other is low (<0.2). A 90-10% split for the training of the ICP and test sets was used to apply training data was further split into proper training and calibration sets using the default Bootstrap Sampler of the nonconformist package. The proper training set was used to train the model, whilst the calibration set is used for calibrating predictions. For this research, we made use of an aggregated conformal predictor (ACP) with 10 ICP models and using mean aggregation[31]. A CP-based model was obtained with a validity of 82% for class 0, the non-permeable class, and 85% for class 1, the permeable class. The conformal metrics are defined for binary classification as:

$$Validity_{class=0} = \frac{\text{Correct single label}_{class=0} + \text{Class "Both" predictions}_{\text{class=0}}}{\text{Number of samples with true label 0}} \quad \text{(Equation 4)}$$

$$Validity_{class=1} = \frac{\text{Correct single label}_{class=1} + \text{Class "Both" predictions}_{\text{class=1}}}{\text{Number of samples with true label 1}} \quad \text{(Equation 5)}$$

$$Efficiency_{class=0} = \frac{\text{Single label predictions}_{class=0}}{\text{Number of samples with true label 0}} \quad \text{(Equation 6)}$$

$$Efficiency_{class=1} = \frac{\text{Single label predictions}_{class=1}}{\text{Number of samples with true label 1}} \quad \text{(Equation 7)}$$

where classes 0 and 1 describe non-permeable and permeable, respectively. Each subscript denoted the true labels of the test set instances.

## Non-conformity score

We used a non-conformity score based on the model probabilities given by the XGBoost model. Figure 7 illustrates how classes are assigned based on a threshold and the distribution of predicted probabilities within the calibration set. The non-conformity score is calculated using the following formula:

$$\alpha = 0.5 - \frac{\hat{P}(y_i|x) - max_{\text{y!=y}_\text{i}}\, \hat{P}(y|x)}{2} \quad \text{(Equation 8)}$$

with non-conformity score $\alpha$, $\hat{P}(y_i|x)$ the probability score for the correct class and $max_{\mathrm{y!=y_i}}\,\hat{P}(y|x)$ the maximum probability of the wrong class(es). A low non-conformity score indicates that the correct class was predicted with high probability, and a high non-conformity score indicates that the incorrect class was predicted with high probability.

*Figure 6: Illustration of the CP-based class assignment at a confidence level of 0.8.*

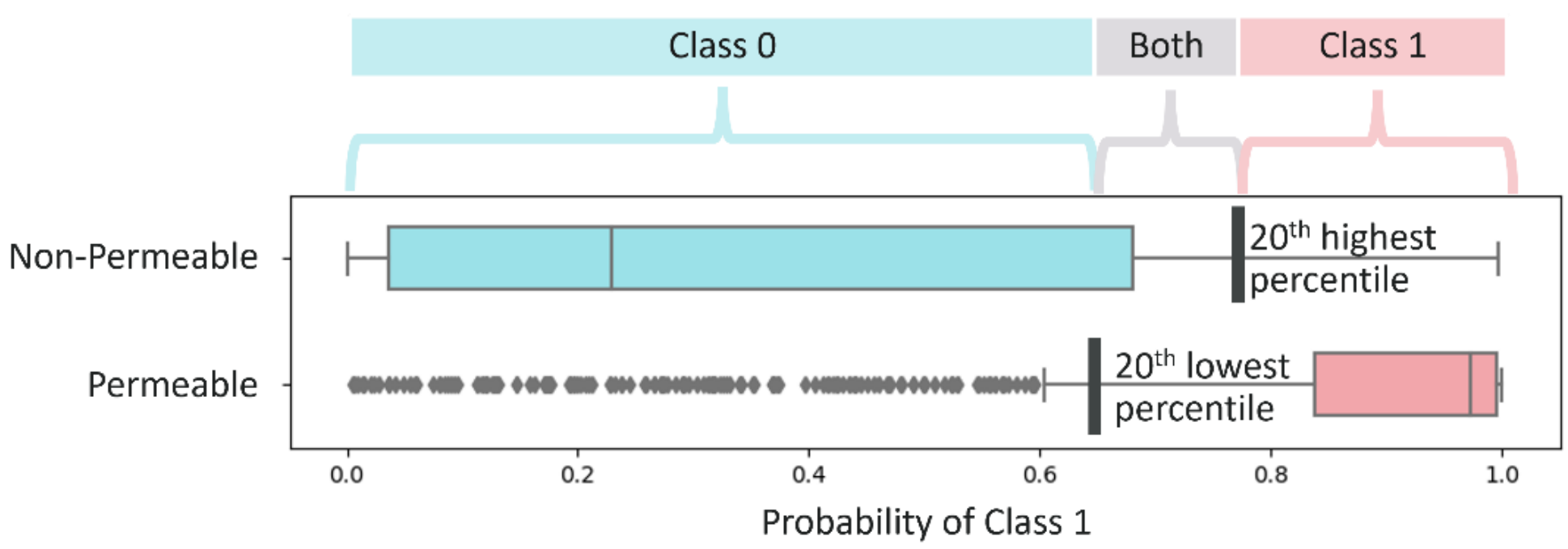

## 5.3. Scoring Functions

Six scoring strategies were explored in this study for integrating the uncertainty of the permeability predictor to the RL loop. The scoring functions that were experimented with include:

I. *RM: Probability Class 1:* This scoring function computes the probability of belonging to the permeable class using the `predict_proba` function of the raw model (RM), XGBoost.

II. *CP: $P_1$ value:* This scoring function calculates the $P_1$ value, representing the p-value of the permeable class of the conformal predictor (CP). Maximizing the $P_1$ value was hypothesized to maximize the confidence that the peptide belongs to the permeable class.

III. *CP: 1-$P_0$ value:* This scoring function calculates the $P_0$ value, representing the p-value of the non-permeable class according to the conformal predictor. Maximizing 1-$P_0$, and consequently minimizing the $P_0$ value, was hypothesized to enhance the confidence of not being non-permeable for a cyclic peptide-like molecule.

IV. *CP: $P_1$-$P_0$ value:* This scoring function aims to maximize the difference between $P_1$ and $P_0$ values. Hence, it increases the confidence of belonging to the permeable class while minimizing the confidence of belonging to the non-permeable class. To map the values from [-1, 1] to [0, 1] range, the following function was used:

$$\text{reinvent}score\ (P_0, P_1) = \frac{(P_1 - P_0) + 1}{2}$$

V. *CP: Harsh function:* This scoring function is aimed at satisfying $P_1$>0.2 and $P_0$<0.2. When both cases are satisfied, the score is assigned a value of 1, otherwise it is given the value of 0. The scoring function aims to optimize the conformal efficiency of the permeable predictions.

$$harsh\ function(P_0, P_1) = \begin{cases} 1, & P_0 \leq 0.2 \wedge P_1 \geq 0.2 \\ 0, & P_0 > 0.2 \vee P_1 < 0.2 \end{cases}$$

VI. *CP: Soft function:* This scoring function aims to satisfy $P_1$> 0.2 and $P_0$<0.2, similarly to the harsh function. However, a step function with three steps is introduced to penalize the two conditions in a more gradual manner compared to the harsh function. When both requirements are met, the scoring function assigns a score of 1. If only one condition is fulfilled, a partial score of 0.5 is given; otherwise, the scoring function yields 0.

$$soft\ function(P_0, P_1) = \begin{cases} 1, & P_0 \leq 0.2 \wedge P_1 \geq 0.2 \\ 0.5, & P_0 \leq 0.2 \vee P_1 \geq 0.2 \\ 0, & P_0 > 0.2 \wedge P_1 < 0.2 \end{cases}$$

No additional transformations were applied to any of the scoring functions, and the RL runs were conducted for 350 epochs with a batch size of 32. To focus on the effect of the scoring function and to eliminate other potential influences, no diversity filter was applied. In addition to the scoring function tested, we also calculated metrics of $P_0$ and $P_1$ from CP during the RL runs, as described above.

## 5.4. Test data

The query for testing the different scoring functions was selected from an external test set, hereby referred to as the Baker dataset[33]. The dataset contains peptides ranging from 6 to 12 amino acids in length, with expanded natural amino acids content. From the Baker dataset, 150 random cyclic peptides with lengths 6, 7 and 10 were selected, reflecting the monomer length distribution that the XGBoost model was trained on. One up to 4 amino acids were randomly masked at random positions for each sequence.

# Supplementary Figures

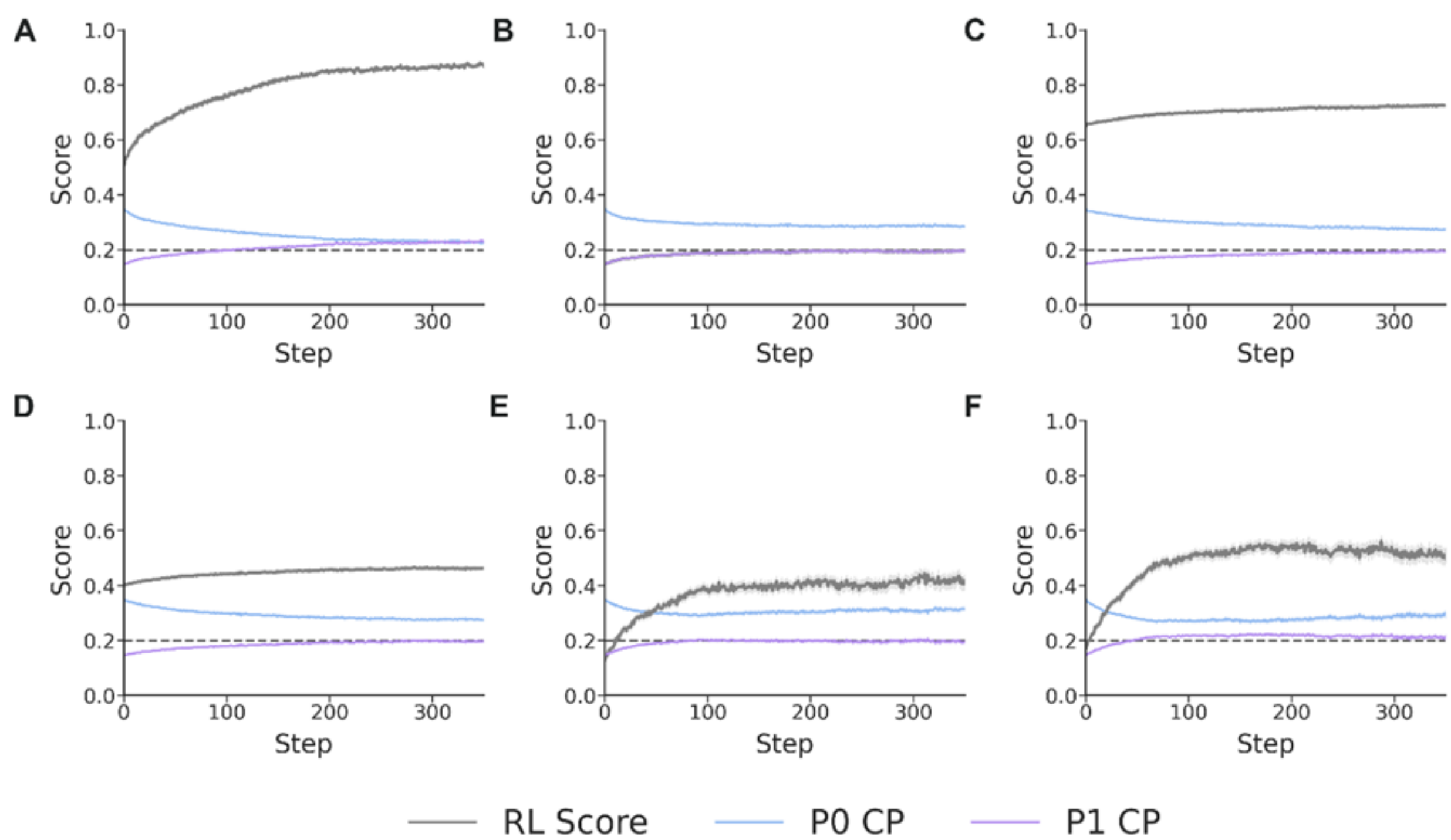

*Figure SI 1: Tracked scores of the experiments using different scoring functions. Average RL Score (grey), CP's $P_0$ (blue) and $P_1$ (purple) values during the RL run. The plots include RL scores (grey) referring to score of the evaluated scoring function, CP's P0 (blue) and $P_1$ (purple) values. Evolution of each scoring component over 350 steps, shown as both the average score and the span across the peptides generated for 150 independent test cases, each performed with a batch size of 32. The scoring function used was (A) the probability of class 1 based on the raw model, (B) $P_1$-value from the CP-based calibrated model, (C) 1- $P_0$ -value from the CP-based calibrated model, (D) $P_1$- $P_0$ value from the CP-based calibrated model, (E) Harsh function based on conformal efficiency, and (F) the soft function. Note that for the $P_1$-value scoring function (B), the $P_1$-value that was tracked overlaps with the RL score since they are the same metric, therefore not visible. Evolutions of each scoring component over 350 steps for the unique valid molecules, are shown as both the average score and the span across the generated peptides for 149 independent test cases.*

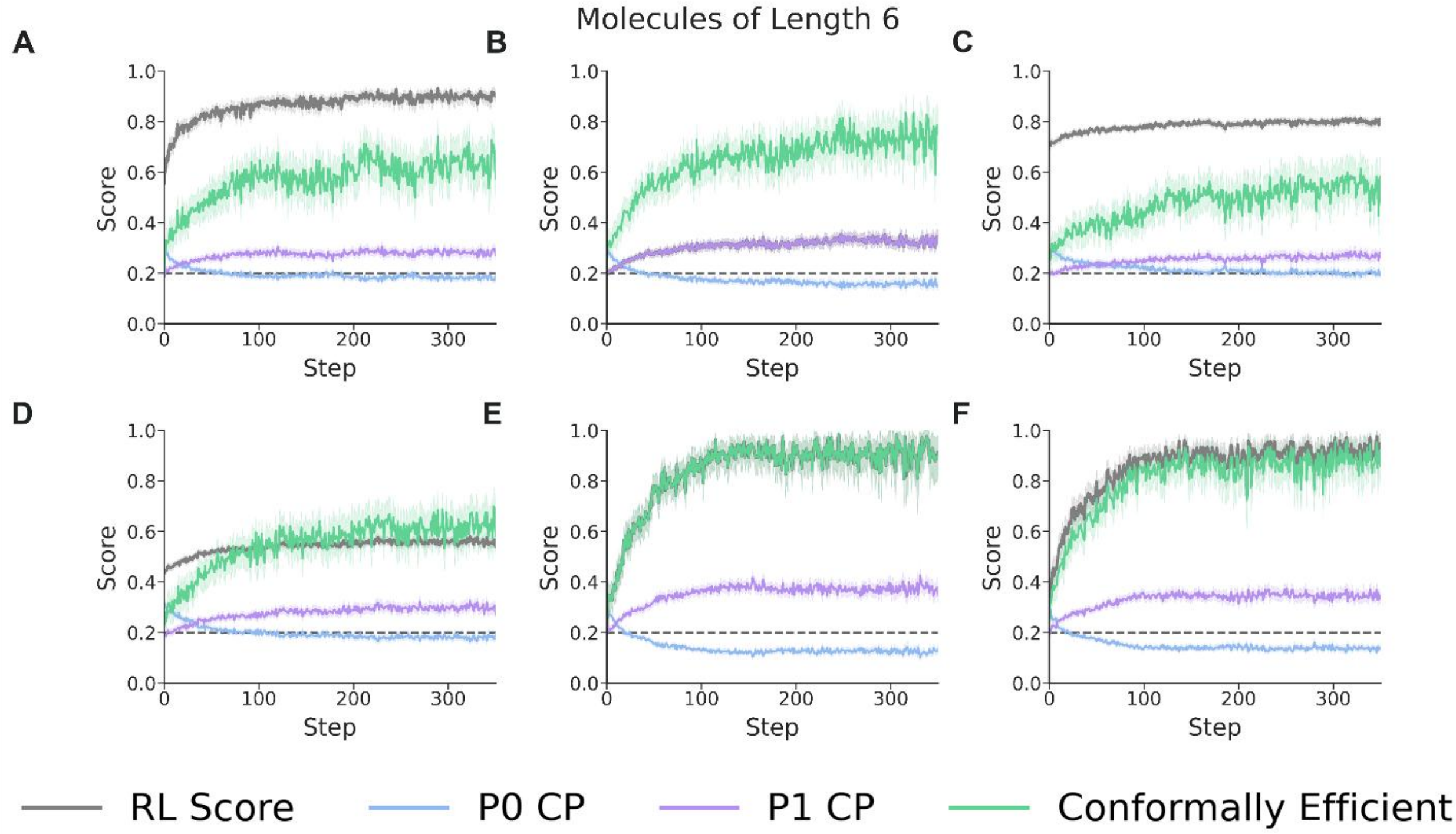

*Figure SI 2: Tracked scores of the experiments using different scoring functions for peptides with length 6. Average RL Score (grey), CP's $P_0$ value (blue), $P_1$ value (purple) and the fraction of conformally efficient molecules predicted to be permeable (green) during the RL run. The plots include RL scores (grey) referring to score of the evaluated scoring function, CP's $P_0$ (blue) and $P_1$ (purple) values. Evolution of each scoring component over 350 steps, shown as both the average score and the span across the peptides generated for 150 independent test cases, each performed with a batch size of 32. The scoring function used was (A) the probability of class 1 based on the raw model, (B) $P_1$-value from the CP-based calibrated model, (C) 1- $P_0$ -value from the CP-based calibrated model, (D) $P_1$- $P_0$ value from the CP-based calibrated model, (E) Harsh function based on conformal efficiency, and (F) the soft function. Note that for the $P_1$-value scoring function (B), the $P_1$-value tracked overlaps with the RL score which is why it is not visible. The same is true for the conformally efficient line in green and the harsh scoring function (E). Evolutions of each scoring component over 350 steps for the unique valid molecules, are shown as both the average score and the span across the generated peptides for 149 independent test cases.*

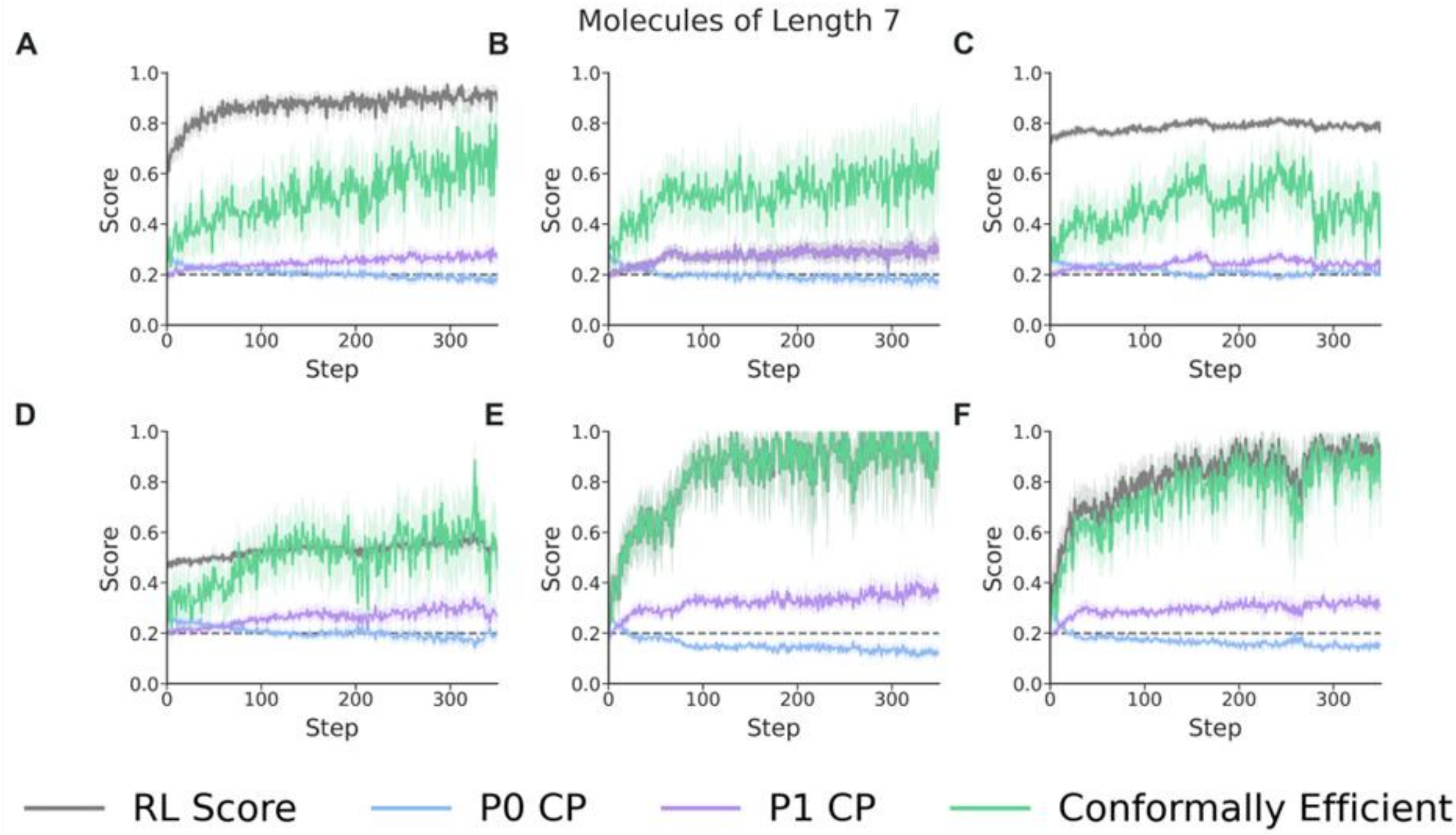

*Figure SI 3: Tracked scores of the experiments using different scoring functions for peptides with length 7. Average RL Score (grey), CP's $P_0$ value (blue), $P_1$ value (purple) and the fraction of conformally efficient molecules predicted to be permeable (green) during the RL run. The plots include RL scores (grey) referring to score of the evaluated scoring function, CP's $P_0$ (blue) and $P_1$ (purple) values. Evolution of each scoring component over 350 steps, shown as both the average score and the span across the peptides generated for 150 independent test cases, each performed with a batch size of 32. The scoring function used was (A) the probability of class 1 based on the raw model, (B) $P_1$-value from the CP-based calibrated model, (C) 1- $P_0$ -value from the CP-based calibrated model, (D) $P_1$- $P_0$ value from the CP-based calibrated model, (E) Harsh function based on conformal efficiency, and (F) the soft function. Note that for the $P_1$-value scoring function (B), the $P_1$-value tracked overlaps with the RL score which is why it is not visible. The same is true for the conformally efficient line in green and the harsh scoring function (E). Evolutions of each scoring component over 350 steps for the unique valid molecules, are shown as both the average score and the span across the generated peptides for 149 independent test cases.*

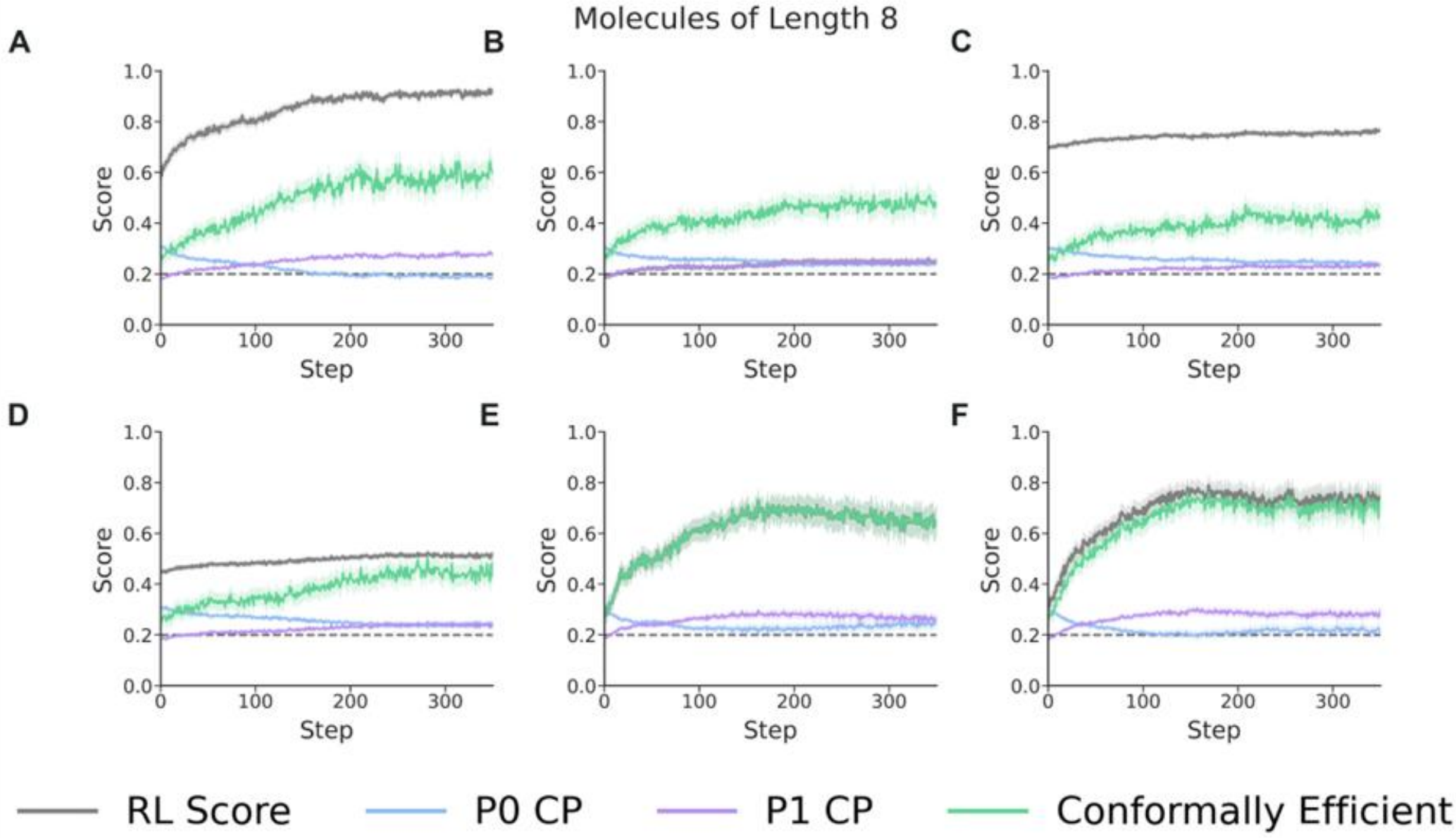

*Figure SI 4: Tracked scores of the experiments using different scoring functions for peptides with length 8. Average RL Score (grey), CP's $P_0$ value (blue), $P_1$ value (purple) and the fraction of conformally efficient molecules predicted to be permeable (green) during the RL run. The plots include RL scores (grey) referring to score of the evaluated scoring function, CP's $P_0$ (blue) and $P_1$ (purple) values. Evolution of each scoring component over 350 steps, shown as both the average score and the span across the peptides generated for 150 independent test cases, each performed with a batch size of 32. The scoring function used was (A) the probability of class 1 based on the raw model, (B) $P_1$-value from the CP-based calibrated model, (C) 1- $P_0$ -value from the CP-based calibrated model, (D) $P_1$- $P_0$ value from the CP-based calibrated model, (E) Harsh function based on conformal efficiency, and (F) the soft function. Note that for the $P_1$-value scoring function (B), the $P_1$-value tracked overlaps with the RL score which is why it is not visible. The same is true for the conformally efficient line in green and the harsh scoring function (E). Evolutions of each scoring component over 350 steps for the unique valid molecules, are shown as both the average score and the span across the generated peptides for 149 independent test cases.*

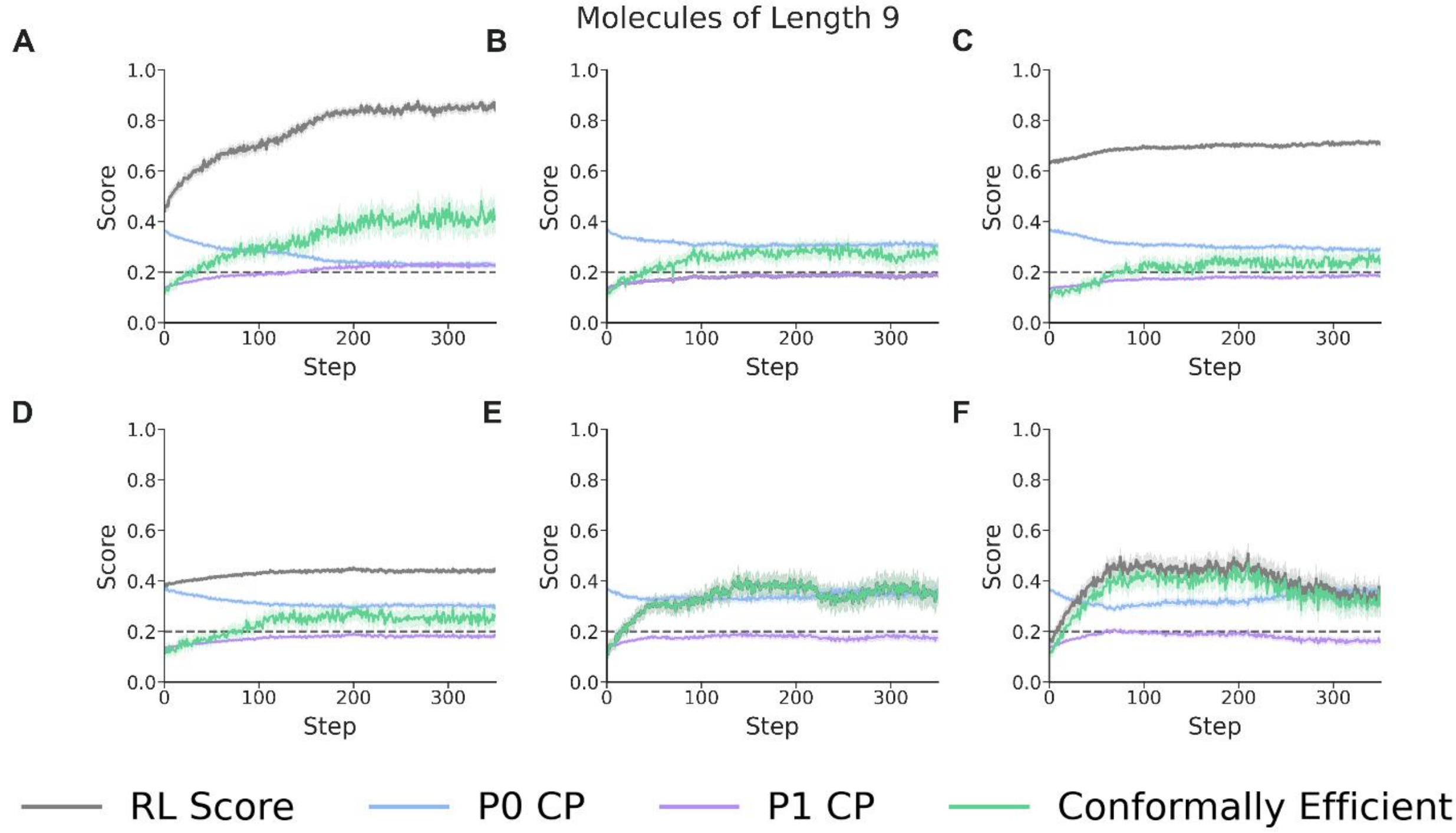

*Figure SI 5: Tracked scores of the experiments using different scoring functions for peptides with length 9. Average RL Score (grey), CP's $P_0$ value (blue), $P_1$ value (purple) and the fraction of conformally efficient molecules predicted to be permeable (green) during the RL run. The plots include RL scores (grey) referring to score of the evaluated scoring function, CP's $P_0$ (blue) and $P_1$ (purple) values. Evolution of each scoring component over 350 steps, shown as both the average score and the span across the peptides generated for 150 independent test cases, each performed with a batch size of 32. The scoring function used was (A) the probability of class 1 based on the raw model, (B) $P_1$-value from the CP-based calibrated model, (C) 1- $P_0$ -value from the CP-based calibrated model, (D) $P_1$- $P_0$ value from the CP-based calibrated model, (E) Harsh function based on conformal efficiency, and (F) the soft function. Note that for the $P_1$-value scoring function (B), the $P_1$-value tracked overlaps with the RL score which is why it is not visible. The same is true for the conformally efficient line in green and the harsh scoring function (E). Evolutions of each scoring component over 350 steps for the unique valid molecules, are shown as both the average score and the span across the generated peptides for 149 independent test cases.*

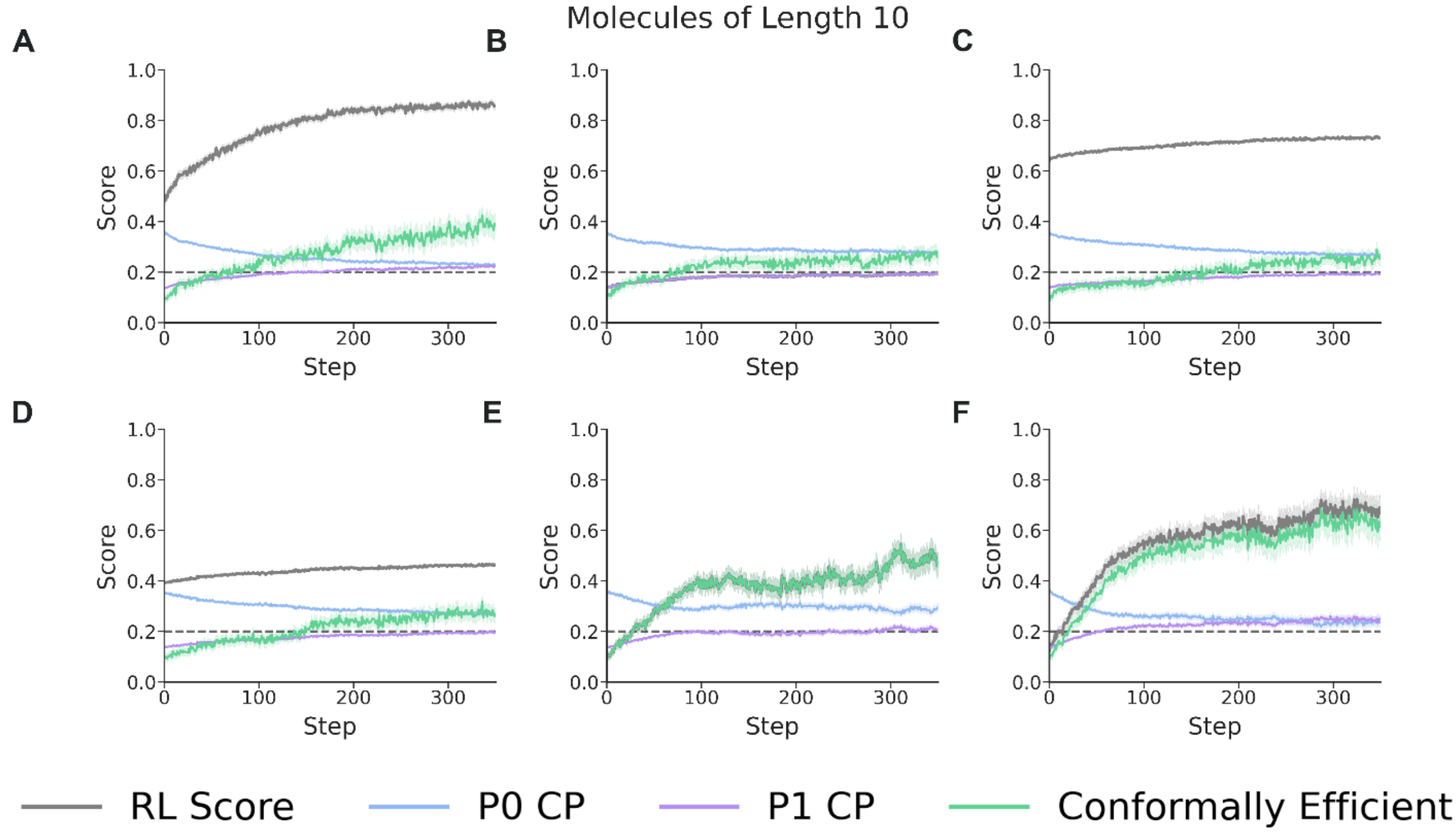

*Figure SI 6: Tracked scores of the experiments using different scoring functions for peptides with length 10. Average RL Score (grey), CP's $P_0$ value (blue), $P_1$ value (purple) and the fraction of conformally efficient molecules predicted to be permeable (green) during the RL run. The plots include RL scores (grey) referring to score of the evaluated scoring function, CP's $P_0$ (blue) and $P_1$ (purple) values. Evolution of each scoring component over 350 steps, shown as both the average score and the span across the peptides generated for 150 independent test cases, each performed with a batch size of 32. The scoring function used was (A) the probability of class 1 based on the raw model, (B) $P_1$-value from the CP-based calibrated model, (C) 1- $P_0$ -value from the CP-based calibrated model, (D) $P_1$- $P_0$ value from the CP-based calibrated model, (E) Harsh function based on conformal efficiency, and (F) the soft function. Note that for the $P_1$-value scoring function (B), the $P_1$-value tracked overlaps with the RL score which is why it is not visible. The same is true for the conformally efficient line in green and the harsh scoring function (E). Evolutions of each scoring component over 350 steps for the unique valid molecules, are shown as both the average score and the span across the generated peptides for 149 independent test cases.*

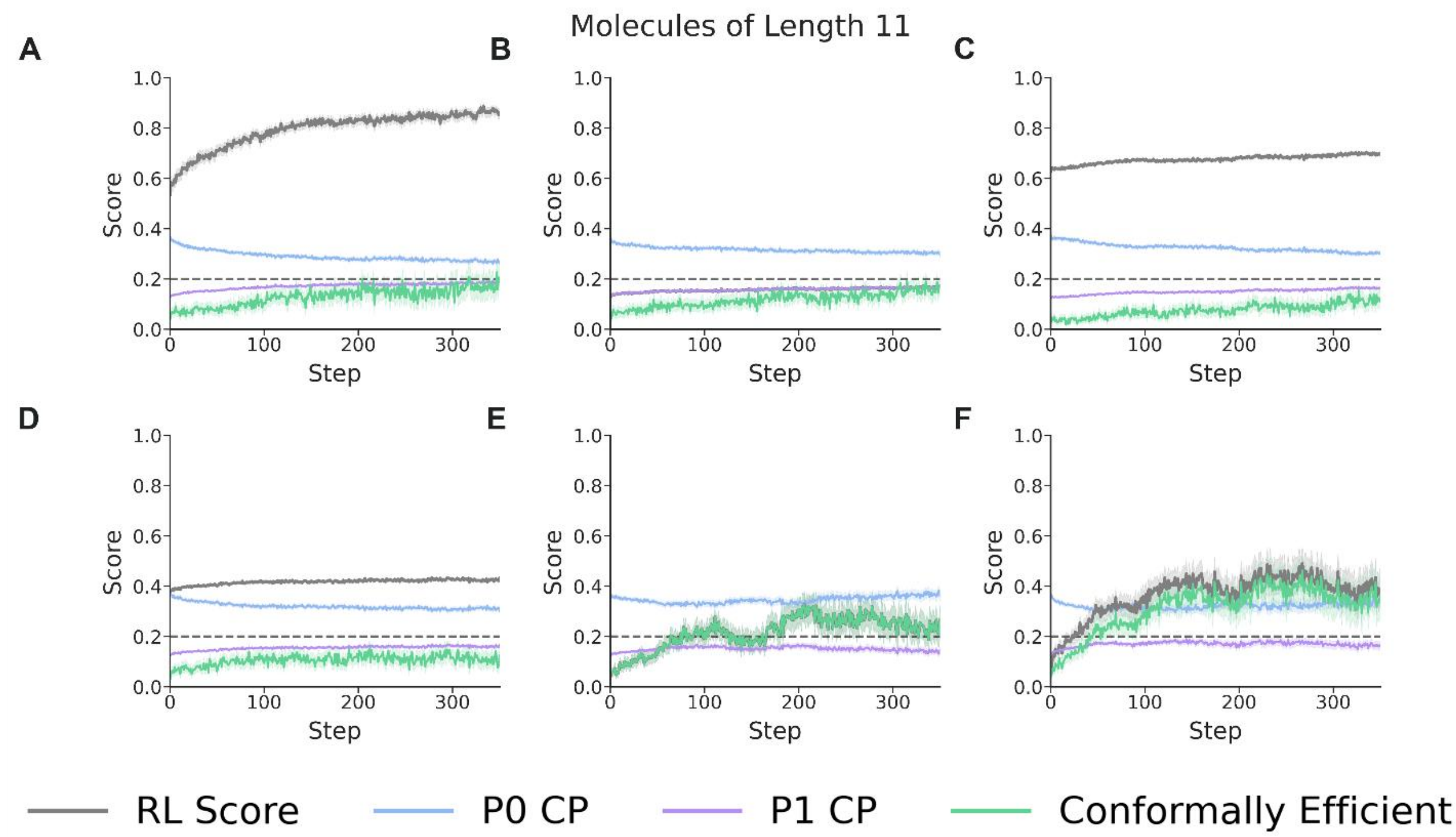

*Figure SI 7: Tracked scores of the experiments using different scoring functions for peptides with length 11. Average RL Score (grey), CP's $P_0$ value (blue), $P_1$ value (purple) and the fraction of conformally efficient molecules predicted to be permeable (green) during the RL run. The plots include RL scores (grey) referring to score of the evaluated scoring function, CP's $P_0$ (blue) and $P_1$ (purple) values. Evolution of each scoring component over 350 steps, shown as both the average score and the span across the peptides generated for 150 independent test cases, each performed with a batch size of 32. The scoring function used was (A) the probability of class 1 based on the raw model, (B) $P_1$-value from the CP-based calibrated model, (C) 1- $P_0$ -value from the CP-based calibrated model, (D) $P_1$- $P_0$ value from the CP-based calibrated model, (E) Harsh function based on conformal efficiency, and (F) the soft function. Note that for the $P_1$-value scoring function (B), the $P_1$-value tracked overlaps with the RL score which is why it is not visible. The same is true for the conformally efficient line in green and the harsh scoring function (E). Evolutions of each scoring component over 350 steps for the unique valid molecules, are shown as both the average score and the span across the generated peptides for 149 independent test cases.*

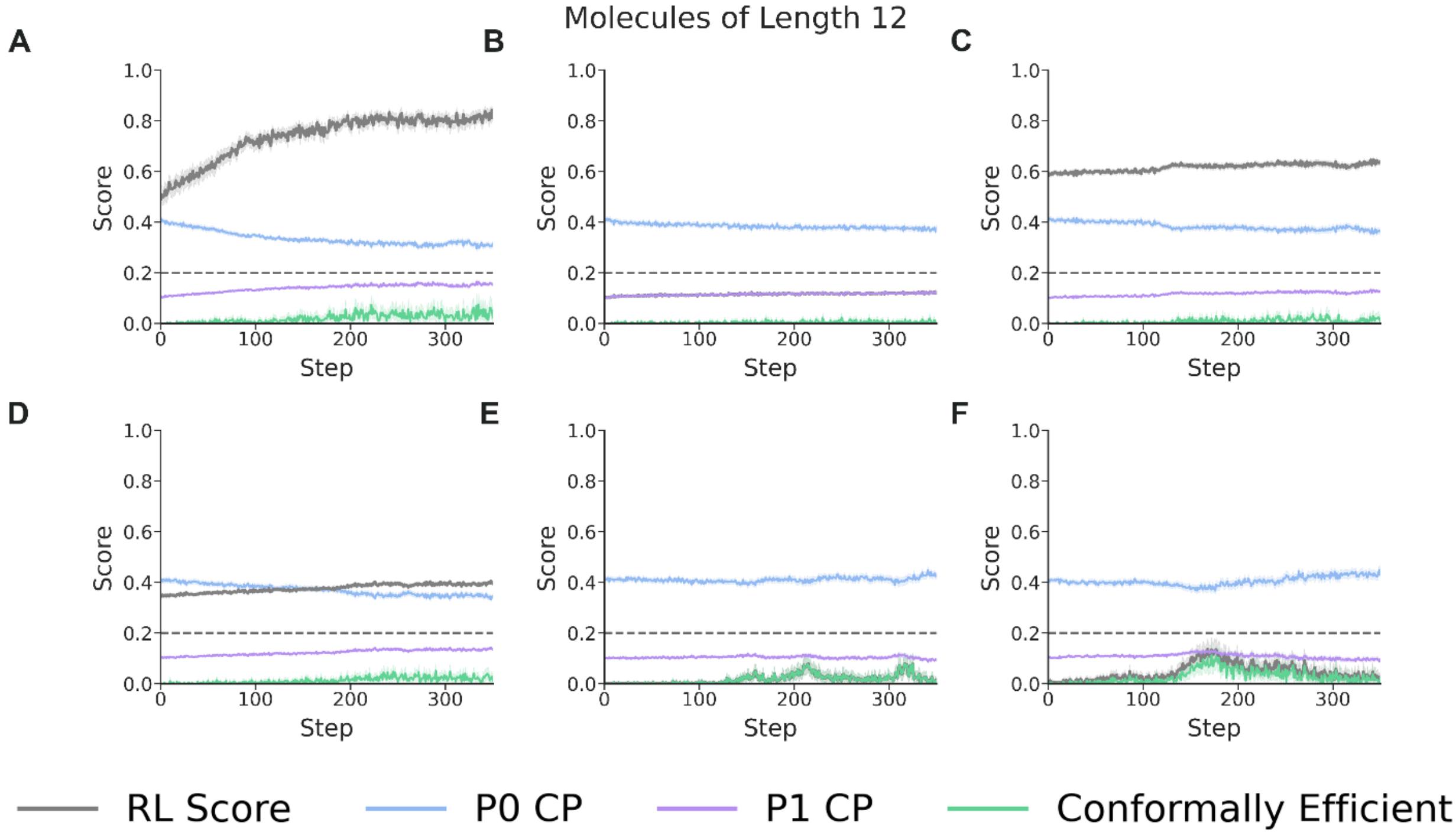

*Figure SI 8: Tracked scores of the experiments using different scoring functions for peptides with length 12. Average RL Score (grey), CP's $P_0$ value (blue), $P_1$ value (purple) and the fraction of conformally efficient molecules predicted to be permeable (green) during the RL run. The plots include RL scores (grey) referring to score of the evaluated scoring function, CP's $P_0$ (blue) and $P_1$ (purple) values. Evolution of each scoring component over 350 steps, shown as both the average score and the span across the peptides generated for 150 independent test cases, each performed with a batch size of 32. The scoring function used was (A) the probability of class 1 based on the raw model, (B) $P_1$-value from the CP-based calibrated model, (C) 1- $P_0$ -value from the CP-based calibrated model, (D) $P_1$- $P_0$ value from the CP-based calibrated model, (E) Harsh function based on conformal efficiency, and (F) the soft function. Note that for the $P_1$-value scoring function (B), the $P_1$-value tracked overlaps with the RL score which is why it is not visible. The same is true for the conformally efficient line in green and the harsh scoring function (E). Evolutions of each scoring component over 350 steps for the unique valid molecules, are shown as both the average score and the span across the generated peptides for 149 independent test cases.*